\documentclass[11pt]{article}
\usepackage{acl2013}
\usepackage{times}
\usepackage{url}
\usepackage{latexsym}
\usepackage{xspace,subfigure,url}
\usepackage[dvipsnames]{color}
\usepackage{multirow}
\usepackage{graphicx}
\usepackage{booktabs}

\newcommand{\word}[1]{\emph{#1}}
\newcommand{\cut}[1]{}

\newcommand{\sectionrule}{\addlinespace[1.5ex]}

\newif{\ifcomment}
\commentfalse

\ifcomment
\newcommand{\cd}[1]{{{\color{blue}{[#1]}}}}
\newcommand{\moritz}[1]{{{\color{cyan}{[#1]}}}}
\newcommand{\dan}[1]{{{\color{Orchid}{[#1]}}}}
\newcommand{\chris}[1]{{{\color{PineGreen}{[#1]}}}}
\newcommand{\jure}[1]{{{\color{red}{[#1]}}}}
\newcommand{\archive}[1]{{{\color{red}{/====#1=====/}}}}

\else
\newcommand{\cd}[1]{}
\newcommand{\moritz}[1]{}
\newcommand{\dan}[1]{}
\newcommand{\chris}[1]{}
\newcommand{\jure}[1]{}
\newcommand{\archive}[1]{}

\fi

\title{A computational approach to politeness with application to social factors}

\cut{
 \author{Cristian Danescu-Niculescu-Mizil$^*$$^\ddagger$, Moritz Sudhof$^\dagger$, Dan Jurafsky$^\dagger$, Jure Leskovec$^*$, \and Christopher Potts$^\dagger$\\ 
$^*$Computer Science Department, $^\dagger$Linguistics Department
\\
 $^*$$^\dagger$Stanford University, $^\ddagger$Max Planck Institute SWS 
\\
\tt \footnotesize
cristiand@cs.stanford.edu, sudhof$\vert$jurafsky@stanford.edu, jure@cs.stanford.edu, cgpotts@stanford.edu}
}

 \author{Cristian Danescu-Niculescu-Mizil$^*$$^\ddagger$, Moritz Sudhof$^\dagger$, Dan Jurafsky$^\dagger$,\\ \textbf{Jure Leskovec$^*$, \and Christopher Potts$^\dagger$}\\ 
$^*$Computer Science Department, $^\dagger$Linguistics Department
\\
 $^*$$^\dagger$Stanford University, $^\ddagger$Max Planck Institute SWS 
\\
\tt \footnotesize
cristiand$\vert$jure@cs.stanford.edu, sudhof$\vert$jurafsky$\vert$cgpotts@stanford.edu}

\begin{document}
\maketitle


\newcommand{\pscore}{Politeness\xspace}
\newcommand{\polscore}{politeness score\xspace}
\newcommand{\topquart}{top quartile\xspace}
\newcommand{\bl}{B\&L\xspace}
\newcommand{\full}{Ling.\xspace}


\begin{abstract}

We propose a computational framework for identifying linguistic
aspects of politeness. Our starting point is a new corpus of requests
annotated for politeness, which we use to evaluate aspects of
politeness theory and to uncover new interactions between politeness
markers and context. 
These findings guide our construction of a classifier with
domain-independent lexical and syntactic features operationalizing key
components of politeness theory, such as \emph{indirection},
\emph{deference}, \emph{impersonalization} and \emph{modality}. Our
classifier achieves close to human performance and is effective across
domains.  We use our framework to study the relationship between
politeness and social power, showing that polite Wikipedia editors are
more likely to achieve high status through elections, but, once elevated,
they become less polite. 
We see a similar negative correlation
between politeness and power on Stack Exchange, where users at the
top of the reputation scale are less polite than those at the bottom. 
Finally, we apply our classifier to a preliminary analysis of  politeness 
variation by gender and
community.

\end{abstract}


\section{Introduction} 
\label{sec:intro}

Politeness is a central force in communication, arguably as basic as
the pressure to be truthful, informative, relevant, and clear
\cite{Grice75,Leech83,Brown:Levinson:1978}. Natural languages provide
numerous and diverse means for encoding politeness and, in
conversation, we constantly make choices about where and how to use
these devices.  \newcite{Kaplan99} observes that ``people desire to be
\emph{paid} respect'' and identifies honorifics and other politeness
markers, like \word{please}, 
 as ``the coin of that payment''.  In turn, politeness markers
are intimately related to the power dynamics of social interactions
and are often a decisive factor in whether those interactions go well
or poorly \cite{GyasiObeng:1997,Chilton:1999,Andersson:Pearson:1999,Rogers:Lee-Wong:2003,Holmes:Stubbe:2005}.

The present paper develops a computational framework for identifying
and characterizing politeness marking in requests. We focus on
requests because they involve the speaker imposing on the addressee,
making them ideal for exploring the social value of politeness
strategies 
\cite{Clark:Schunk:1980,Francik:Clark:1985}. Requests also stimulate extensive use of what
\newcite{Brown:PolitenessSomeUniversalsInLanguageUsage:1987} call
\emph{negative politeness}: speaker strategies for minimizing (or
appearing to minimize) the imposition on the addressee, for example,
by being indirect (\word{Would you mind}) or apologizing for the
imposition (\word{I'm terribly sorry, but})
\cite{Lakoff73,RLakoff:1977,Brown:Levinson:1978}. 

Our investigation is guided by a new corpus of requests annotated for
politeness.
    The data come from two large online communities in
which members frequently make requests of other members: Wikipedia,
where the requests involve editing and other administrative functions,
and Stack Exchange, where the requests center around a diverse range
of topics (e.g., programming, gardening, cycling).  The corpus
confirms the broad outlines of linguistic theories of politeness
pioneered by
\newcite{Brown:PolitenessSomeUniversalsInLanguageUsage:1987}, but it
also reveals new interactions between politeness markings and the
morphosyntactic context.  For example, the politeness of \word{please}
depends on its syntactic position and the politeness markers it
co-occurs with.

Using this corpus, we construct a politeness classifier with a wide
range of domain-independent lexical, sentiment, 
and
dependency features operationalizing key components of politeness
theory, including not only the negative politeness markers mentioned
above but also elements of \emph{positive politeness} 
(gratitude, positive and
optimistic sentiment, solidarity, and inclusiveness). 
The classifier achieves near human-level accuracy across domains,
which highlights the consistent nature of politeness strategies and
paves the way to using the classifier to study new data.

Politeness theory predicts a negative correlation between politeness
and the power of the requester, where power is broadly construed to
include social status, authority, and autonomy \cite{Brown:PolitenessSomeUniversalsInLanguageUsage:1987}.  The greater the
speaker's power 
relative to
her addressee, the less polite her
requests are expected to be: there is no need for her to incur the
expense of paying respect, and failing to make such payments can
invoke, and hence reinforce, her power. We support this prediction by
applying our politeness framework to Wikipedia and Stack Exchange,
both of which provide independent measures of social status.  We show
that polite Wikipedia editors are more likely to achieve high status
through elections; however, once elected, they become less polite. Similarly, on Stack Exchange, we find that users at the
top of the reputation scale are less polite than those at the bottom. 

Finally, we briefly address the question of how politeness norms vary
across communities and social groups.  Our findings confirm
established results about the relationship between politeness and
gender, and they identify substantial variation in politeness across
different programming language subcommunities on Stack Exchange.


\section{Politeness data} 
\label{sec:data}

Requests involve an imposition on the addressee, making them a natural
domain for studying the inter-connections between linguistic aspects
of politeness and social variables.

\paragraph{Requests in online communities} 
\label{sub:request_data}

We base our analysis on two online communities where requests have
an important role: the Wikipedia community of editors
and the Stack
Exchange question-answer community.\footnote{\url{http://stackexchange.com/about}} 
On Wikipedia, to coordinate
on the creation and maintenance of the collaborative encyclopedia,
editors can interact with each other on user
talk-pages;\footnote{\url{http://en.wikipedia.org/wiki/Wikipedia:User_pages}}
requests posted on a user talk-page, although public, are generally directed to the owner of the talk-page.
On Stack Exchange, users
often comment on existing posts requesting further information or
proposing edits; these requests are generally directed to the authors
of the original posts. 

Both communities are not only rich in user-to-user requests, but
these requests are also part of consequential conversations, not empty social banter; they solicit specific information or concrete
actions, and they expect a response.

\cut{
In this study, we focus on two online communities where users make 
 requests of other users: Wikipedia and Stack Exchange. On Wikipedia, to facilitate the curation of the collaborative encyclopedia, editors can ask other editors direct questions on ``talk pages." On Stack Exchange, a collection of question-answer communities, the formal questions themselves are not directed at any particular user, but users often ask direct questions at specific users in the comments on a post by asking for clarification or additional information.
}

\cut{
Our data comprises a total of 35,661 user talk-page requests from Wikipedia and 373,519 comment requests from multiple Stack Exchange subcommunities (programming, physics, cooking, etc.) along with metadata, which will be discussed in context.
 }

\paragraph{Politeness annotation} 
\label{sub:politeness_annotation}

Computational studies of politeness, or indeed any  aspect of linguistic pragmatics, demand richly labeled data.
We therefore label a large portion of our request data (over 10,000 utterances) using Amazon Mechanical Turk (AMT), 
creating the largest corpus with politeness annotations (see Table~\ref{tab:data} for details).\footnote{Publicly available at \url{http://www.mpi-sws.org/~cristian/Politeness.html}
}

We choose to annotate requests containing exactly two sentences,
where the second sentence is the actual request (and ends with a
question mark).  This provides enough context to the annotators while
also controlling for length effects.  Each annotator
was instructed to read a batch of 13 requests and consider them as
originating from a co-worker by email.  For each request, the annotator
had to indicate how polite she perceived the request to be by using
a slider with values ranging from ``very impolite'' to ``very
polite''.\footnote{We used non-categorical ratings for finer granularity and to help
account for annotators' different perception scales.}
Each request was labeled by five different annotators.

We vetted annotators by restricting their residence to be in the U.S.\ and by conducting a
linguistic background questionnaire.  We also gave them
a paraphrasing task shown to be effective for 
verifying and eliciting  
 linguistic attentiveness
\cite{munro2010crowdsourcing}, and we monitored the annotation job
and manually filtered out annotators who submitted uniform or
seemingly random annotations. 

\begin{figure}[t]
\centering
\includegraphics[width=0.47\textwidth]{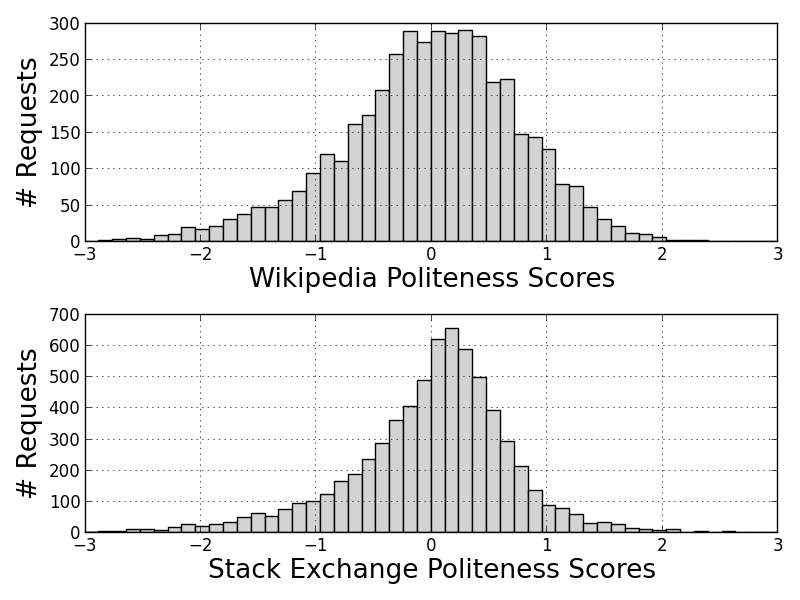}
  \caption{Distribution of politeness scores. Positive scores indicate requests perceived as polite.}  
  \label{fig:annotation}
\end{figure}

Because politeness is highly subjective and annotators may have inconsistent scales, we applied the standard z-score normalization to each worker's scores.
Finally, we define the politeness score (henceforth \emph{politeness}) of a request as the average of the five scores assigned by the annotators. The distribution of resulting request scores (shown in Figure 
\ref{fig:annotation}) has an average of 0 and standard deviation of 0.7 
 for both domains; positive values correspond to polite requests (i.e., requests with normalized annotations towards the ``very polite'' extreme) and negative values to impolite requests.  A summary of all our request data is shown in Table~\ref{tab:data}.

\begin{table}[t]
\begin{tabular}{l@{\hspace{5pt}}c@{ \hspace{5pt}}c@{ \hspace{5pt}}c}

\toprule
\textbf{domain} & \textbf{\#requests} & \textbf{\#annotated} & \textbf{\#annotators} \\
\midrule
Wiki &  35,661 & 4,353 & 219\\
SE & 373,519& 6,604 & 212\\
\bottomrule
\end{tabular}
\caption{Summary of the request data and its politeness annotations.}
\label{tab:data}
\end{table}

\paragraph{Inter-annotator agreement} 
\label{par:inter_annotator_agreement}

\begin{figure}[t]
\centering
\vspace*{-12pt}
\includegraphics[width=0.47\textwidth]{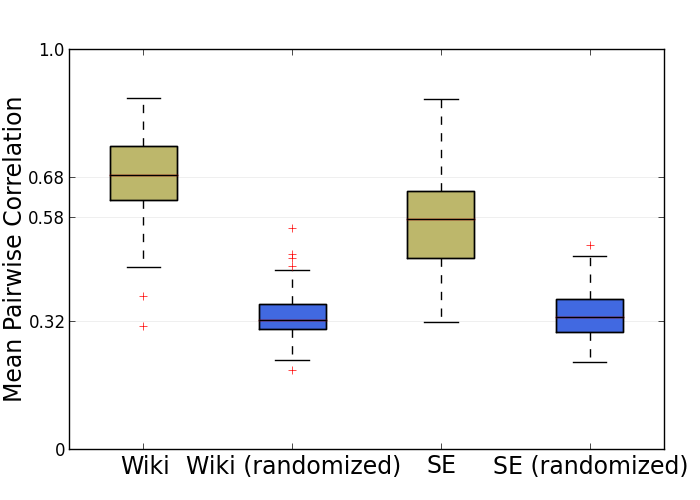}
  \caption{Inter-annotator pairwise correlation, compared to the same measure after randomizing the scores.
  \cut{According to the Wilcoxon test, human-based correlation vectors and randomly-generated correlation vectors were significantly different, with pvalues of 2.07e-21 (Wikipedia) and 1.76e-51 (Stack Exchange).} }  
  \label{fig:interannotator}
\end{figure}

To evaluate the reliability of the annotations we measure the inter-annotator agreement by computing, for each batch of 13 documents that were annotated by the same set of 5 users, the mean pairwise correlation of the respective scores.  For reference, we compute the same quantities after randomizing the scores by sampling from the observed distribution of politeness scores.  As shown in Figure~\ref{fig:interannotator}, the labels are coherent and significantly different from the randomized procedure ($p<0.0001$ according to a Wilcoxon signed rank test).\footnote{The commonly used Cohen/Fleiss Kappa agreement measures are not suitable for this type of annotation, in which labels are continuous rather than categorical.} 

\paragraph{Binary perception} 
\label{par:polite_vs_impolite}
Although we did not impose a discrete categorization of politeness, we acknowledge an implicit binary perception of the phenomenon: whenever an annotator moved a slider in one direction or the other, she made a binary politeness judgment. 
However,
 the boundary between somewhat polite and somewhat impolite requests  
can be blurry.
To test this intuition, we break the set of annotated requests into four groups, each corresponding to a politeness score quartile.  For each quartile, we compute the percentage of requests for which all five annotators made the same binary politeness judgment. As shown in Table~\ref{tab:interagreement},  full agreement is much more common in the 
1\textsuperscript{st} (bottom) and 4\textsuperscript{th} (top) quartiles than in the middle quartiles. This 
 suggests that the politeness scores assigned to requests that are only somewhat polite or somewhat impolite are less reliable and less tied to an intuitive notion of binary politeness. This discrepancy motivates our choice of classes in the prediction experiments (Section~\ref{sec:prediction}) and our use of the top politeness quartile (the 25\% most polite requests) as a reference in our subsequent discussion.

\begin{table}[t]
\begin{center}
\begin{tabular}{ ccccl}
\toprule
 \textbf{Quartile:} & 1\textsuperscript{st}&2\textsuperscript{nd} &3\textsuperscript{rd}&4\textsuperscript{th}\\
 \midrule

{Wiki}&62\%&8\%&3\%&51\%\\
{SE} & 37\%&4\%&6\%&46\%\\
\bottomrule

\end{tabular}
\end{center}

\caption{The percentage of requests for which all five annotators agree on binary politeness. The 4\textsuperscript{th} quartile contains the requests with the top 25\% politeness scores in the data. (For reference, randomized scoring yields agreement percentages of \textless $20\%$ for all quartiles.)
\cut{Probability that all workers agree on the fundamental politeness of a document, by quartile.}}
\label{tab:interagreement}
\end{table}


\section{Politeness strategies} 
\label{sec:experiments}

As we mentioned earlier, 
requests
impose on the addressee, potentially placing her in social peril if
she is unwilling or unable to comply.
Requests therefore naturally give rise to
the negative politeness strategies of
\newcite{Brown:PolitenessSomeUniversalsInLanguageUsage:1987},
which are attempts to mitigate these social threats. These strategies are
prominent in Table~\ref{tab:strategies}, which describes the core
politeness markers we 
analyzed in our corpus of Wikipedia requests. 
We do not include the Stack Exchange data in this analysis,  reserving it as a ``test community'' 
for our prediction task (Section~\ref{sec:prediction}).

\cut{
In the
table, the politeness scores are those described in
section~\ref{sub:politeness_annotation} and summarized in
figure~\ref{fig:politeness-scores}.  They range from roughly $-4$ to
$+4$. The annotations are normally distributed with a mean of $0$
(s.d.~XXXX). Negative scores correspond to less polite expressions,
positive scores to more polite ones.

\cd{Quartile discussion}  correlated, more intuive  
The top quartiles
}

Requests exhibiting politeness markers are automatically extracted
using regular expression matching on the dependency parse obtained
by the Stanford Dependency Parser \cite{deMarneffe06}, together with
specialized lexicons.
For example, for the hedges marker (Table~\ref{tab:strategies}, line~\ref{f:hedges}), we match all requests
containing a nominal subject dependency edge pointing out from a
hedge verb from the hedge list created by \newcite{hyland:05}.  For each politeness strategy,
Table~\ref{tab:strategies} shows the average \textbf{politeness}
score of the respective requests (as described in
Section~\ref{sub:politeness_annotation}; positive numbers indicate
polite requests), and their \textbf{top politeness quartile}
membership (i.e., what percentage fall 
within
 the top quartile of
politeness scores). 
As discussed at the end of Section~\ref{sec:data}, 
the top politeness quartile gives a more robust
and more intuitive measure of politeness.  For
reference, a random sample of requests will have a 0 politeness
score and a 25\% top quartile membership; in both cases, larger
numbers indicate higher politeness.

Gratitude and deference
(lines~\ref{f:gratitude}--\ref{f:deference}) are ways for the
speaker to 
incur a social cost,  helping to balance out the
burden the request places on the addressee.  Adopting
\newcite{Kaplan99}'s metaphor, these are the coin of the realm when it
comes to paying the addressee respect.  Thus, they are 
indicators of positive politeness.

Terms from the sentiment lexicon  \cite{liu+al:05}
are also tools for positive politeness, either
by emphasizing a positive relationship with the addressee 
(line~\ref{f:poslexicon}), or
being impolite by using negative sentiment that damages this positive relationship 
(line~\ref{f:neglexicon}).
Greetings (line~\ref{f:indirectgreet}) are another way to build a positive relationship with the addressee.

The remainder of the cues in  Table~\ref{tab:strategies}
are 
negative politeness strategies,
serving the purpose of minimizing, at least in appearance, the imposition on the addressee.
Apologizing
(line~\ref{f:apologizing}) deflects the social threat of the request by attuning to the imposition itself.
Being indirect (line~\ref{f:indirectbtw}) is
another way to minimize social threat.  
This strategy allows the speaker to
avoid words and phrases conventionally associated
with requests. 
First-person plural forms like \word{we} and \word{our}
(line~\ref{f:1pl}) are also ways of being indirect, as they create the
sense that the burden of the request is shared between speaker and
addressee (\word{We really should \ldots}).  Though indirectness is
not invariably interpreted as politeness marking
\cite{Blum-Kulka:2003}, it is nonetheless a reliable marker of it, as
our scores indicate.  What's more, direct variants (imperatives,
statements about the addressee's obligations) are less polite
(lines~\ref{f:direct}--\ref{f:directstart}).

Indirect strategies also combine with hedges (line~\ref{f:hedges})
conveying that the addressee is unlikely to accept the burden
(\word{Would you by any chance \ldots?}, \word{Would it be at all
  possible \ldots?}).  These too serve to provide the addressee with a
face-saving way to deny the request.  We even see subtle effects of
modality at work here: the irrealis, counterfactual forms \word{would}
and \word{could} are more polite than their ability (dispositional) or
future-oriented variants \word{can} and \word{will}; compare
lines~\ref{f:could} and \ref{f:can}.  This parallels the contrast
between factuality markers (impolite; line~\ref{f:factuality}) and
hedging (polite; line~\ref{f:hedges}).

Many of these features are correlated with each other, in keeping with
the insight of
\newcite{Brown:PolitenessSomeUniversalsInLanguageUsage:1987} that
politeness markers are often combined to create a cumulative effect of
increased politeness.  Our corpora also highlight interactions that
are unexpected (or at least unaccounted for) on existing theories of
politeness.  For example, sentence-medial \word{please} is polite
(line~\ref{f:please}), presumably because of its freedom to combine
with other negative politeness strategies (\word{Could you please
  \ldots}). In contrast, sentence-initial \word{please} is impolite
(line~\ref{f:pleasestart}), because it typically signals a more direct
strategy (\word{Please do this}), which can make the politeness marker
itself seem insincere. We see similar interactions between pronominal
forms and syntactic structure: sentence-initial \word{you} is impolite
(\word{You need to \ldots}),
 whereas sentence-medial \word{you} is
often part of the indirect strategies we discussed above
(\word{Would/Could you \ldots}).

\newcommand{\signone}{\textsuperscript{}}
\newcommand{\sigone}{\textsuperscript{*}}
\newcommand{\sigtwo}{\textsuperscript{**}}
\newcommand{\sigthree}{\textsuperscript{***}}

\newcounter{fcounter}

\setcounter{fcounter}{0}
\newcommand{\fitem}[1]{\refstepcounter{fcounter}\thefcounter.\label{#1}}

\begin{table*}[ht]
  \begin{center}
    \setlength{\tabcolsep}{2pt}
    \begin{tabular}{r @{ } l  r@{}l @{\hspace{25pt}} r@{}l l}
      \toprule
                              & \textbf{Strategy}   & \multicolumn{2}{l}{\textbf{\pscore}} & \multicolumn{2}{c}{\textbf{In \topquart}} & \multicolumn{0}{c}{ \textbf{Example}}\\
      \midrule
      \sectionrule
      \fitem{f:gratitude}     & Gratitude            &     0.87&\sigthree  & \hspace{15pt} 78\%&\sigthree & \textbf{I} really \textbf{appreciate} that you've done them.\\ 
      \fitem{f:deference}     & Deference            &    0.78&\sigthree  & 70\%&\sigthree & \textbf{Nice work} so far on your rewrite. \\ 
            \fitem{f:indirectgreet} & Greeting  &    0.43&\sigthree  & 45\%&\sigthree & \textbf{Hey}, I just tried to \ldots \\ 
      \fitem{f:poslexicon} & Positive lexicon & 0.12&\sigthree & 32\%&\sigthree &  \textbf{Wow}! / This is a \textbf{great} way to deal\ldots\\
      \fitem{f:neglexicon} & Negative lexicon & -0.13&\sigthree & 22\%&\sigtwo   & If you're going to \textbf{accuse} me \ldots \\
            \sectionrule
      \fitem{f:apologizing}   & Apologizing          &    0.36&\sigthree  & 53\%&\sigthree & \textbf{Sorry} to bother you \ldots \\ 
      \sectionrule
      \fitem{f:please}        & Please               &    0.49&\sigthree  & 57\%&\sigthree & Could you \textbf{please} say more\ldots \\
      \fitem{f:pleasestart}   & Please start         & $-$0.30&\sigone    & 22\%&\signone  & \textbf{Please} do not remove warnings \ldots \\ %
      \sectionrule
      \fitem{f:indirectbtw}   & Indirect (btw)       &    0.63&\sigthree  & 58\%&\sigtwo &\textbf{By the way}, where did you find \ldots \\
      \fitem{f:direct}        & Direct question      & $-$0.27&\sigthree  & 15\%&\sigthree & \textbf{What} is your native language? \\ 
      \fitem{f:directstart}   & Direct start         & $-$0.43&\sigthree  &  9\%&\sigthree & \textbf{So} can you retrieve it or not? \\
      \sectionrule
      \fitem{f:could}         & Counterfactual modal  &    0.47&\sigthree  & 52\%&\sigthree  & \textbf{Could}/\textbf{Would} you \ldots \\
      \fitem{f:can}           & Indicative modal      &    0.09&\signone   & 27\%&\signone   & \textbf{Can}/\textbf{Will} you \ldots \\
      \sectionrule
      \fitem{f:1start}        & 1st person start     &    0.12&\sigthree  & 29\%&\sigtwo &\textbf{I} have just put the article \ldots \\  
      \fitem{f:1pl}           & 1st person pl.       &    0.08&\sigone    & 27\%&\signone & Could \textbf{we} find a less complex name \ldots \\
    \fitem{f:1}             & 1st person           &    0.08&\sigthree  & 28\%&\sigthree & It is \textbf{my} view that ... \\ 
      \fitem{f:2}             & 2nd person           &    0.05&\sigthree  & 30\%&\sigthree & But what's the good source \textbf{you} have in mind? \\ 
      \fitem{f:2start}        & 2nd person start     & $-$0.30&\sigthree  & 17\%&\sigtwo & \textbf{You}'ve reverted yourself \ldots \\ 
      \sectionrule
      \fitem{f:hedges}        & Hedges               & 0.14&\sigthree  & 28\%&\signone & I \textbf{suggest} we start with \ldots \\ 
      \fitem{f:factuality}    & Factuality          & $-$0.38&\sigthree & 13\%&\sigthree & \textbf{In fact} you did link, \ldots \\
      \bottomrule
    \end{tabular}
    \caption{Positive (\ref{f:gratitude}-\ref{f:neglexicon})  and negative (\ref{f:apologizing}--\ref{f:factuality}) politeness strategies
    and their relation to human perception of politeness.
    For each strategy we show the average (human annotated) \textbf{politeness} scores for the requests exhibiting that strategy (compare with 0 for a random sample of requests; a positive number indicates the strategy is perceived as being polite), as well as the percentage of requests exhibiting the respective strategy that fall in the \textbf{top quartile} of politeness scores (compare with 25\% for a random sample of requests). Throughout the paper: for politeness scores, statistical significance is calculated by comparing the set of requests exhibiting the strategy with the rest using a Mann-Whitney-Wilcoxon U test; for top quartile membership a binomial test is used.}
    \label{tab:strategies}
  \end{center}
\end{table*}


\section{Predicting politeness} 
\label{sec:prediction}

We now show how our linguistic analysis can be used in a machine learning model 
for automatically classifying requests according to politeness.  
A classifier can help verify the predictive power, robustness, and domain-independent generality of
the linguistic strategies of Section~\ref{sec:experiments}.
Also, by providing automatic politeness judgments for large amounts of new data on a scale unfeasible for human annotation,
it can also enable a detailed analysis of the relation between politeness and
social factors (Section~\ref{sec:correlates}).

\paragraph{Task setup} 
\label{par:task_setup}

To evaluate the robustness and domain-independence of the analysis from  Section~\ref{sec:experiments},
we run our prediction experiments on two
very different domains. We treat Wikipedia as a
``development domain'' since we used it for developing
and identifying features and for training our models. 
Stack Exchange is our ``test domain'' since it was not used for identifying features. 
We take 
the
model  (features and weights) trained on Wikipedia and use them to classify requests from Stack Exchange.

We consider two classes of requests:
\textbf{polite}  and \textbf{impolite}, defined as the top and,
respectively, bottom quartile of requests when sorted by their
 politeness
  score (based on the binary notion of politeness 
  discussed in
   Section~\ref{sec:data}).
The classes are therefore balanced, with each class
consisting of 1,089 requests for the Wikipedia domain and 1,651
requests for the Stack Exchange domain.

We compare two classifiers --- a  bag of words classifier (BOW) and a linguistically informed classifier (\full) --- and use human labelers as a reference point.
The  BOW classifier is an SVM 
using a unigram feature representation.\footnote{Unigrams appearing less than 10 times are excluded.} 
We consider this to be a strong baseline for this new classification task, especially considering the large amount of training data available. 
The linguistically informed classifier (\full) is an SVM using the linguistic features listed in Table~\ref{tab:strategies} in addition to the unigram features. 
Finally, 
to obtain a reference point for the prediction task we also collect three new politeness annotations for each of the requests in our dataset using the same methodology described in Section~\ref{sec:data}. 
We then calculate human performance on the task (Human) as the percentage of requests for which the average score from the additional annotations matches the binary politeness class of the original annotations (e.g., a positive score corresponds to the polite class).

\paragraph{Classification results} 
\label{par:classification_results}
We evaluate the classifiers both in an in-domain setting, with a standard leave-one-out cross validation procedure, and in a cross-domain setting, where we train on one domain and test on the other (Table~\ref{tab:prediction}).
For both our development and our test domains, and in both the in-domain and
cross-domain settings, the linguistically informed features 
give 3-4\% absolute improvement over the
bag of words model.  While the in-domain
results are within 3\% of human performance, 
the greater room for improvement in the cross-domain setting motivates
further research on linguistic cues of politeness.

The experiments in this section confirm that our
theory-inspired features are indeed effective in practice, and
generalize well to new domains.
In the next section we exploit this insight to automatically annotate
a much larger set of requests 
(about 400,000)
 with politeness labels,
enabling us to relate politeness to several social variables and
outcomes. For new requests, we use class probability estimates
obtained by fitting a logistic regression model to the output of
the SVM \cite{witten2005data} as {\em predicted politeness scores}
(with values between 0 and 1; henceforth {\em politeness}, by abuse of language).

\begin{table}[t]
\begin{tabular}{l@{\hspace{5pt}}cc@{\hspace{10pt}}cc}
&\multicolumn{2}{c}{In-domain } &\multicolumn{2}{c}{Cross-domain} \\
Train & Wiki & SE &Wiki  &SE\\
Test & Wiki & SE &SE& Wiki \\ \midrule
BOW & 79.84\%& 74.47\% & 64.23\% & 72.17\%\\
\full & {83.79\%} & {78.19\%} & {67.53\%} & {75.43\%}\\
\midrule
Human & 86.72\% & 80.89\%&80.89\%&86.72\%\\
\end{tabular}
\caption{Accuracies of our two classifiers for Wikipedia (Wiki) and Stack Exchange (SE), for in-domain and cross-domain settings.  Human performance is included as a reference point.  The random baseline performance is 50\%.}
\label{tab:prediction}
\end{table}


\section{Relation to social factors} 
\label{sec:correlates}

We now apply our framework to studying the relationship between
politeness and social variables, focussing on social power dynamics.
Encouraged by the close-to-human performance of our in-domain
classifiers, we use them to assign politeness labels to our full
dataset and then compare these labels to independent measures of power
and status in our data. The results closely match those obtained with
human-labeled data alone, thereby supporting the use of computational
methods to pursue questions about social variables.

\subsection{Relation to social outcome} 
\label{sub:status}

Earlier, we characterized politeness markings as currency used to pay
respect. Such language is therefore costly in a social sense, and,
relatedly, tends to incur costs in terms of communicative efficiency
\cite{van2003being}. Are these costs worth paying? We now address this
question by studying politeness in the context of the electoral system
of the Wikipedia community of editors.

Among Wikipedia editors, status is a salient social variable \cite{anderson2012effects}. 
Administrators (\emph{admins}) are editors who have been granted
certain rights, including the ability to block other editors and to
protect or delete
articles.\footnote{\url{http://en.wikipedia.org/wiki/Wikipedia:Administrators}}
Admins have a higher status than common editors (\emph{non-admins}),
and this distinction seems to be widely acknowledged by the community
\cite{Burke:2008:TUM:1358628.1358871,Leskovec+Huttenlocher+Kleinberg:2010a,Danescu-Niculescu-Mizil+al:12a}.
Aspiring editors become admins through public
elections,\footnote{\url{http://en.wikipedia.org/wiki/Wikipedia:Requests_for_adminship}}
so we know when the status change from non-admin to admins occurred
and can study users' language use in relation to that time.

To see whether politeness correlates with eventual high status, we
compare, in Table~\ref{tab:status}, the politeness levels of requests
made by users who will eventually succeed in becoming administrators
(Eventual status: Admins) with requests made by users who are not
admins (Non-admins).\footnote{We consider only requests made up to one month before the election, to avoid confusion with pre-election behavior.} 
 We observe that admins-to-be are significantly more polite than non-admins. 
One might wonder whether this merely
reflects the fact that not all users aspire to become admins, and
those that do are more polite. To address this, we also consider users
who ran for adminship but did not earn community approval (Eventual
status: Failed). These users are also significantly less polite than
their successful counterparts, indicating that politeness indeed
correlates with a positive social outcome here.

\begin{table}[tb]
	\begin{center}
		\begin{tabular}{l @{\hspace{5pt}}r@{}lr@{}l}
		\toprule
		\textbf{Eventual status} & \multicolumn{2}{c}{\textbf{Politeness}}& \multicolumn{2}{c}{\textbf{Top quart.}}\\
		\midrule
			Admins & \hspace{15pt}0.46&\sigtwo & \hspace{10pt} 30\%&\sigthree\\
			Non-admins   & 0.39&\sigthree &  25\%&\\
			Failed & 0.37&\sigtwo &  22\%&\\		
		\bottomrule
		\end{tabular}
	\end{center}
	\caption{Politeness and status. 
          Editors who will eventually become admins are more polite than non-admins 
          (p$<$0.001 according to a  Mann-Whitney-Wilcoxon U test) and than editors 
           who will eventually fail
            to become admins (p$<$0.001).  Out of their requests, 
          30\% are rated in the top politeness quartile (significantly more than the 
          25\% of a random sample; p$<$0.001 according to a binomial test). 
          This analysis was conducted on 
          31k requests 
           (1.4k 
           for
            Admins, 
          28.9k for Non-admins, 
          652 for Failed).
        }
	\label{tab:status}
      \end{table}

\begin{figure}[tb]
	\begin{center}
		\includegraphics[width=0.45\textwidth]{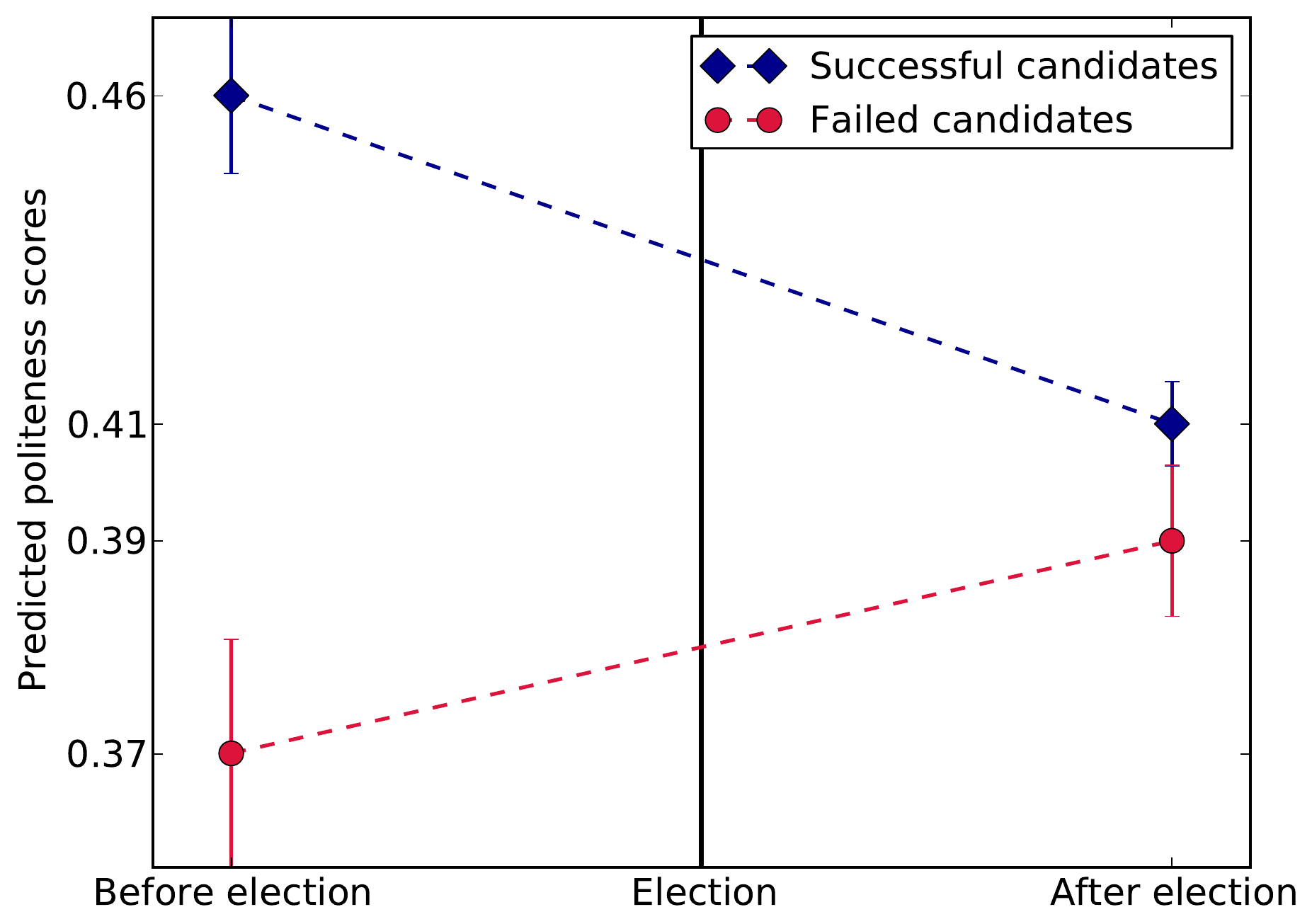}
	\end{center}
	\caption{Successful and failed  candidates before and after elections.  Editors that will eventually succeed (diamond marker) are significantly more polite than those that will fail (circle markers). Following the elections, successful editors become less polite while unsuccessful editors become more polite.}
	\label{fig:status}
\end{figure}

\subsection{Politeness and power} 
\label{sub:relation_to_role}

We expect a rise in status to correlate with a decline in politeness (as predicted by politeness theory,
and
discussed in Section~\ref{sec:intro}).
The previous section does not test this hypothesis, since all editors
compared in Table~\ref{tab:status} had the same (non-admin) status
when writing the requests. However, our data does provide three ways of
testing this hypothesis.

First, after the adminship elections, successful editors get a boost
in power by receiving admin privileges. Figure~\ref{fig:status} shows
that this boost is mirrored by a significant decrease in politeness
(blue, diamond markers). Losing an election has the opposite effect
on politeness (red, circle markers), perhaps as a consequence of
reinforced low status.

Second, Stack Exchange allows us to test more situational 
power
effects.\footnote{We restrict all experiments in this section to the largest subcommunity of Stack Exchange, namely Stack Overflow.} 
On the site, users request, from the community, information
they are lacking. This informational asymmetry between the question-asker
and his audience puts him at a social disadvantage. We therefore
expect the question-asker to be more polite than the people who respond.
Table~\ref{tab:role} shows that this expectation is born out: comments
posted to a thread by the original question-asker are more polite than those
posted by other users. 

\begin{table}[htb]
	\begin{center}
		\begin{tabular}{l @{\hspace{5pt}}r@{}lr@{}l}
		\toprule
		\textbf{Role} & \multicolumn{2}{c}{\textbf{Politeness}}& \multicolumn{2}{c}{\textbf{Top quart.}}\\
		\midrule
			Question-asker  & \hspace{15pt}0.65&\sigthree & \hspace{10pt} 32\%&\sigthree\\
			Answer-givers    & 0.52&\sigthree &  20\%&\sigthree\\
		\bottomrule
		\end{tabular}
	\end{center}
	\caption{Politeness and dependence. Requests made in comments posted by the question-asker are significantly more polite than the other requests. Analysis conducted on 
		181k
	 requests 
	(106k for question-askers, 75k for answer-givers).
	}\label{tab:role}

\end{table}

Third, Stack Exchange allows us to examine power in the form of
authority, through the community's reputation system. Again, we see a
negative correlation between politeness and power, even after
controlling for the role of the user making the requests
(i.e., Question-asker or ~Answer-giver).
Table~\ref{tab:reputation} summarizes the results.\footnote{Since our
  data does not contain time stamps for reputation scores, we only
  consider requests that were issued in the
  six months prior to the available snapshot.} 
\begin{table}[h]
	\begin{center}
		\begin{tabular}{l @{\hspace{5pt}}r@{}lr@{}l}
		\toprule
		\textbf{Reputation level} & \multicolumn{2}{c}{\textbf{Politeness}}& \multicolumn{2}{c}{\textbf{Top quart.}}\\
		\midrule
			 Low reputation   & \hspace{15pt}0.68 & \sigthree & \hspace{10pt} 27\% & \sigthree\\ 
			Middle reputation &       		 0.66        & \sigthree &        25\%        & \\ 
			 High reputation  &       		 0.64        & \sigthree &        23\%        & \sigthree\\ 
		\bottomrule
		\end{tabular}
	\end{center}
	\caption{Politeness and Stack Exchange reputation (texts by question-askers only). 
          High-reputation users are less polite. Analysis conducted on 25k requests
          (4.5k low, 12.5k middle, 8.4k high). }
	\label{tab:reputation}
\end{table}

\paragraph{Human validation} 
\label{par:human_validation}
The above analyses are based on predicted politeness from our
classifier. This allows us to use the entire request data corpus to
test our hypotheses and to apply precise controls to our experiments
(such as restricting our analysis to question-askers in the reputation
experiment). 
In order to validate 
this methodology, we turned again to human annotation: we collected additional politeness annotation for the types of requests involved in the newly designed experiments. 
When we re-ran our experiments on human-labeled data alone we obtained
the same qualitative
results, with statistical significance always lower than 0.01.\footnote{However, due to the limited size of the human-labeled data, we could not control for the role of the user in the Stack Exchange reputation experiment.}

\cut{
\begin{table}[tb]
	\begin{center}
		\begin{tabular}{l@{\hspace{25pt}}r@{}lr@{}l}
		\toprule
		\textbf{PL name} & \multicolumn{2}{c}{\textbf{Politeness}} & \multicolumn{2}{l}{\textbf{Top quartile}} \\
		\midrule
C	     	      &     0.45 &   \sigthree          &  \hspace{15pt}  22\%   &\sigone           \\
C++	 	      &         0.47 &  \sigthree       &        23\%   &\sigone       \\
Python	     &          0.47 & \sigthree       &         23\%         \\
Java	 	     &      0.51 &           &     24\%             \\
Perl	 	     &      0.49 &           &     24\%             \\
Php	  	     &          0.51 &       &         24\%         \\
C\#	  	     &          0.52 &\sigtwo        &         25\%         \\
Javascript     &        0.53 &\sigtwo         &       26\%   &\sigtwo        \\
Objective-C     &       0.56 &\sigtwo          &      28\%   &\sigtwo         \\
Ruby		     &      0.59 &\sigthree            &     28\%    &\sigone         \\
		\bottomrule
		\end{tabular}
	\end{center}
	\caption{Politeness of requests from different programming language communities on Stack Exchange. (Analysis conducted on 27.5k requests.)\moritz{FYI: request \#s: [(u'c', 1149), (u'java', 4443), (u'c#', 6606), (u'javascript', 3493), (u'c++', 3040), (u'perl', 382), (u'python', 2071), (u'objective-c', 966), (u'php', 4707), (u'ruby', 579)]}}
	\label{tab:pl}
\end{table}
}

\paragraph{Prediction-based interactions} 
\label{par:other_interactions}

\cut{
This suggests that our framework could be further used to explore differences in politeness levels across factors of interest, such as communities, geographical regions and gender where annotating sufficient data becomes infeasible (at least in a hypothesis refinement phase).  For example, we  find highly significant evidence that Ruby, Objective-C and Javascript programmers the politest among the programming languages communities, with C, C++ and Python at the other end of the politeness spectrum\footnote{Ironically, this analysis was conducted mostly in Python.}.  Wikipedians from the U.S. Midwest are most polite (when comparing with other census-defined regions), and female wikipedians are generally more polite. 
}

The human validation of classifier-based results suggests that our
prediction framework can be used to explore differences in politeness
levels across factors of interest, such as communities, geographical
regions and gender, even where gathering sufficient human-annotated
data is infeasible. We mention just a few such preliminary results
here: (i) Wikipedians from the U.S.~Midwest are most polite (when
compared to other census-defined regions), (ii) female Wikipedians
are generally more polite (consistent with prior studies in which
women are more polite in a variety of domains;
\cite{herring1994politeness}), and (iii) programming language
communities on Stack Exchange vary significantly by politeness
(Table~\ref{tab:pl}; full disclosure: our
analyses were conducted in Python).

\begin{table}[tb]
  \begin{center}
    \begin{tabular}{l@{\hspace{25pt}}r@{}lr@{}l}
      \toprule
      \textbf{PL name} & \multicolumn{2}{c}{\textbf{Politeness}} & \multicolumn{2}{l}{\textbf{Top quartile}} \\
      \midrule
      Python	     &          0.47 & \sigthree       &         23\%         \\
      Perl	 	     &      0.49 &           &     24\%             \\
      PHP	  	     &          0.51 &       &         24\%         \\
      Javascript     &        0.53 &\sigtwo         &       26\%   &\sigtwo        \\
      Ruby		     &      0.59 &\sigthree            &     28\%    &\sigone         \\
      \bottomrule
    \end{tabular}
  \end{center}
  \caption{Politeness of requests from different 
  language communities on Stack Exchange.}
  \label{tab:pl}
\end{table}

\section{Related work} 
\label{sec:relwork}

Politeness has been a central concern of modern pragmatic theory since
its inception
\cite{Grice75,Lakoff73,RLakoff:1977,Leech83,Brown:Levinson:1978},
because it is a source of pragmatic enrichment, social meaning, and
cultural variation
\cite{Harada76,Matsumoto:1988,Ide:1989,Blum-Kulka:Kasper:1990,Blum-Kulka:2003,Watts:2003,Byon:2006}.
The starting point for most research is the theory of
\newcite{Brown:PolitenessSomeUniversalsInLanguageUsage:1987}.
Aspects of this theory have been explored from game-theoretic
perspectives \cite{van2003being} and implemented in language
generation systems for interactive narratives
\cite{Walker:1997:ILS:267658.267680}, cooking instructions,
\cite{gupta2007rude}, translation \cite{FarPad:2012}, spoken dialog
\cite{WanFinOga:2012}, and subjectivity analysis
\cite{Abdul-Mageed:Diab:2012}, among others.

In recent years, politeness has been studied in online settings. 
Researchers have
identified variation in politeness marking across different contexts
and media types
\cite{herring1994politeness,Brennan:1999:WEC:295666.295942,Duthler:JournalOfComputerMediatedCommunication:2006}
and between different social groups \cite{burke2008mind}. The present
paper pursues similar goals using orders of magnitude more data, which facilitates a
fuller survey of different politeness strategies.

Politeness marking is one aspect of the broader issue of how language
relates to power and status, which has been studied in the context of
workplace discourse
\cite{Bramsen+al:2011a,Diehl+Namata+Getoor:2007a,PetHohXia:2011,prabhakaran2012predicting,Gilbert:ProceedingsOfCscw:2012,McCallum+Wang+CorradaEmmanuel:2007a}
and social networking \cite{scholand2010social}.
 However, this
research focusses on domain-specific textual cues, whereas the present
work seeks to leverage domain-independent politeness cues, building on
the literature on how politeness affects worksplace social dynamics
and power structures
\cite{GyasiObeng:1997,Chilton:1999,Andersson:Pearson:1999,Rogers:Lee-Wong:2003,Holmes:Stubbe:2005}.
\newcite{Burke:2008:TUM:1358628.1358871} 
study
 the question of how
and why specific individuals rise to administrative positions on
Wikipedia, and \newcite{Danescu-Niculescu-Mizil+al:12a} show that
power differences on Wikipedia are revealed through aspects of
linguistic accommodation.  The present paper complements this work by
revealing the role of politeness in social outcomes and power relations.


\section{Conclusion} 
\label{sec:conclusion}

We construct and release a large collection of politeness-annotated requests and use it to evaluate key aspects of politeness theory.  We build a politeness classifier that achieves near-human performance and use it to explore the relation between politeness and social factors such as power, status, gender, and community membership. We hope the publicly available collection of annotated requests enables further study of politeness and its relation to social factors, as this paper has only begun to explore this area.

\section*{Acknowledgments}

We thank Jean Wu for running the AMT annotation task, and all the participating turkers. 
We thank Diana Minculescu and the anonymous reviewers for their helpful comments.
This work was supported in part by NSF
IIS-1016909,        
CNS-1010921,        
IIS-1149837,        
IIS-1159679,        
ARO MURI,           
DARPA SMISC,        
Okawa Foundation,   
Docomo,             
Boeing,             
Allyes,             
Volkswagen,         
Intel,              
Alfred P. Sloan Fellowship,    
the Microsoft Faculty Fellowship, 
the Gordon and Dailey Pattee Faculty Fellowship,
and the Center for Advanced Study in the Behavioral Sciences at Stanford.

\end{document}